\documentclass[times,final]{elsarticle}

\usepackage[english]{babel}
\usepackage{comment}

\usepackage[letterpaper,top=2cm,bottom=2cm,left=3cm,right=3cm,marginparwidth=1.75cm]{geometry}

\usepackage{amsmath}
\usepackage{graphicx}
\usepackage[colorlinks=true, allcolors=blue]{hyperref}
\usepackage[table]{xcolor}
\usepackage{tikz}
\usepackage{adjustbox} 
\usepackage{booktabs, multirow} 
\usepackage{soul}
\usepackage{changepage,threeparttable} 

\begin{document}

\begin{frontmatter}
\title{A reinforcement learning strategy to automate and accelerate h/p-multigrid solvers} 

\author[1]{David Huergo\corref{cor1}}
\cortext[cor1]{Corresponding author}
\ead{david.huergo.perea@upm.es}
\author[1]{Laura Alonso}
\author[1]{Saumitra Joshi}
\author[1]{Adrian Juanicotena}
\author[1,2]{Gonzalo Rubio}
\author[1,2]{Esteban Ferrer}

\address[1]{ETSIAE-UPM-School of Aeronautics, Universidad Politécnica de Madrid, Plaza Cardenal Cisneros 3, E-28040 Madrid, Spain}
\address[2]{Center for Computational Simulation, Universidad Politécnica de Madrid, Campus de Montegancedo, Boadilla del Monte, 28660 Madrid, Spain}


\begin{abstract}
We explore a reinforcement learning strategy to automate and accelerate h/p-multigrid methods in high-order solvers. 
Multigrid methods are very efficient but require fine-tuning of numerical parameters, such as the number of smoothing sweeps per level and the correction fraction (i.e., proportion of the corrected solution that is transferred from a coarser grid to a finer grid). 
The objective of this paper is to use a proximal policy optimization algorithm to automatically tune the multigrid parameters and, by doing so, improve stability and efficiency of the h/p-multigrid strategy. 
%

Our findings reveal that the proposed reinforcement learning h/p-multigrid approach significantly accelerates and improves the robustness of steady-state simulations for one dimensional advection-diffusion and nonlinear Burgers' equations, when discretized using high-order h/p methods, on uniform and nonuniform grids. 
\end{abstract}

\end{frontmatter}


\section{Introduction}

Multigrid methods are widely recognized for minimizing time-to-convergence \cite{wesseling2001,stueben2001} in numerical solvers and have become an essential tool also in the family of high-order (HO) methods \cite{fehn2020}. These methods exploit the fact that errors represented on coarser discrete spaces have higher spatial frequencies than on the original discrete space, enabling faster damping. Traditional multigrid methods, proposed in the 1970s \cite{brandt1977}, rely on successively coarser meshes and are thus termed \textit{h-multigrid}. 
In the context of HO solvers, we can represent high-order errors on lower orders, effectively coarsening the polynomial order ($P$). The resulting \textit{p-multigrid} offers simplicity in transfer operations between levels, making it a natural choice to accelerate HO methods. \textit{H/p-multigrid} techniques have been extensively employed to solve  elliptic, Euler, and the compressible Navier-Stokes equations \cite{haga2011,luo2012,ghidoni2014,franciolini2020, ruedaramirez2019,botti2017,krank2017,fehn2018,joshi2023length}. A recent comprehensive summary by Fehn et al. covers relevant publications on \textit{h}- and \textit{p-multigrid} \cite{hp-multigrid}. To extract maximum benefits from the multigrid in the context of HO methods, a combination of \textit{h-} and \textit{p-multigrid} is advisable, as demonstrated in studies exploring \textit{h/p-multigrid} for compressible laminar flows \cite{NASTASE2006330,shahbazi2009}.

Among HO methods, the Flux Reconstruction (FR) approach, introduced by Huynh \cite{huynh2007}, has gained popularity due to its flexibility and precision \cite{WITHERDEN2016227}. FR exhibits high accuracy with spectral polynomial convergence, a compact stencil, and the ability to handle unstructured grids. Unifies various existing HO methods (including popular discontinuous Galerkin variants) under a common framework, facilitating the discovery of new approaches \cite{vincent2011}. In particular, it simplifies implementation by avoiding the need for surface or volume integrals, using the differential form of the governing equations \cite{huynh2014}. This paper explores the performance of the FR method for steady-state simulations, where multigrid methods can significantly accelerate convergence.

Multigrid methods, while very effective, present their own set of challenges. They involve numerous parameters and the performance of the method heavily depends on the correct selection of these parameters. This selection process can be complex and time-consuming, often requiring expert knowledge or extensive trial and error.
In light of these challenges, this paper proposes the construction of an expert system based on reinforcement learning (RL) to automate the selection of parameters in the multigrid method. The objective is to enhance efficiency and accuracy when calculating the steady-state solution for advection-diffusion and Burgers' equations. This approach could potentially eliminate the need for manual parameter tuning, thereby saving time and reducing the reliance on expert intervention.



In RL, which belongs to a branch of machine learning known as semi-supervised learning, an agent learns the optimal behavior, also known as optimal policy, by interacting with an environment to maximize cumulative rewards over time \cite{DBLP:books/lib/SuttonB98}. 
%
The environment, that is the FR solver, provides a piece of information, known as state, to the RL agent, which takes an action consequently. During the training process, for each state-action pair, 
the agent receives a feedback in the form of a reward that lets the agent understand whether it is performing well or not. By facing different situations, the agent can converge to the optimal policy.

This optimal policy can be obtained through two main distinct approaches: value-based methods and policy-based methods.
In value-based methods, agents estimate a value function that predicts expected rewards for each state. Then, by selecting actions that lead to future states with the highest expected values, agents adhere to an optimal policy.
In contrast, policy-based methods directly optimize the policy without first estimating a value function. This approach offers several advantages over value-based methods, including better handling of high-dimensional action spaces, applicability to continuous and discrete action environments (while value-based methods struggle with continuous actions), and typically smoother convergence properties (although they can still converge to local minima) \cite{garnier2021review,viquerat2021direct,dussauge2023reinforcement,huergo2024reinforcement}. 
Here, we employ the Proximal Policy Optimization (PPO) algorithm \cite{PPO}, which is a policy-based algorithm that has gained significant attention for its effectiveness and stability in solving complex optimization problems \cite{rabault2020deep}. It belongs to the family of policy gradient methods \cite{PolGrad}, which employ a gradient ascent algorithm to estimate the weights of the optimal policy by interacting with an environment. 
In the context of multigrid methods, PPO can be leveraged to accelerate convergence and enhance the efficiency of the flux-reconstruction solver.

In summary, this work presents the integration of RL (i.e. PPO) with multigrid to automate the intricate selection of parameters associated with multigrid algorithms. We aim to significantly enhance computational speed and efficiency (without losing stability) when addressing advection-diffusion equations. In particular, we focus on scenarios where these equations are discretized using high-order flux-reconstruction methods. To our knowledge, this work is the first that aims to control and improve multigrid methods using RL.

The remainder of the paper is organized as follows. First, we detail the methodology, including FR and RL, to then present the results for linear advection-diffusion and Burgers' equations. We finish with some conclusions and outlooks. 

\section{Methodology}

\subsection{Advection-diffusion and nonlinear Burgers' equations}



In this work, we concentrate on the one-dimensional linear advection-diffusion (\ref{eq: ad-dif}) and Burgers' (\ref{eq: burgers}) equations:

\begin{equation}
\label{eq: ad-dif}
 \begin{gathered}
    \frac{\partial u}{\partial t} + a \frac{\partial u}{\partial x} = \nu \frac{\partial^2 u}{\partial x^2}, 
    \end{gathered}
\end{equation}

\begin{equation}
\label{eq: burgers}
\begin{gathered}
\frac{\partial u}{\partial t}+u \frac{\partial u}{\partial x}=\nu \frac{\partial^2 u}{\partial x^2} + f,\\
\end{gathered}
\end{equation}
\newline
where $\nu \in \mathbb{R}$ is the kinematic viscosity and $a \in \mathbb{R}$ is the constant advection velocity, $u(x, t): \mathbb{R} \times \mathbb{R} \rightarrow \mathbb{R}$ is the scalar quantity of interest, $t \in \mathbb{R}^{+}$ is the time and  $x \in[0,1]$ the spatial coordinate. We impose the initial condition: 
\begin{equation}
u(x,0) = \frac{1}{4}\left(\sin{2\pi x} + \sin{3\pi x} + \sin{4\pi x} + \sin{6\pi x} + \sin{8\pi x} + \sin{10\pi x} + \sin{12\pi x}\right),
\end{equation}
along with Dirichlet boundary conditions matching the exact solution of the problem. 
The source term, $f$, added to the Burgers' equation so that the steady-state solution is not constant, is defined as:
\begin{equation}
    f(x)= 2 \pi \left(\cos^2{2 \pi x}-\sin^2{2 \pi x} \right) + \nu 2 \pi \left(\sin{2 \pi x} + \cos{2 \pi x}\right).
\end{equation}





\subsection{Flux reconstruction method}

To derive a discrete FR form for the advection-diffusion and Burgers' equations, we consider the general case of a 1D scalar conservation law:

\begin{equation}
\label{1D scalar conservation law}
\frac{\partial u}{\partial t}+\frac{\partial f(u)}{\partial x}=0,
\end{equation}
being $x$ the spatial coordinate of the solution points, $t$ the time, $u=u(x, t)$ is a conserved scalar quantity and $f=f(u)$ is the flux, defined on a space $\Omega$. The flux is $f=au$ for advection-diffusion and $f=1/2u^2$ for Burgers' equation.

We discretize $\Omega$ into $N$ different cells, $\Omega_j=\left\{x \mid x_j<x<x_{j+1}\right\}$.
With the domain partitioned, we can approximate the exact solution $u$ in eq. (\ref{1D scalar conservation law}) using numerical solutions $u_j^\delta$, where $\delta$ denotes the discrete nature of the terms. 
These solutions are defined as polynomials of degree $P$ within each $\Omega_j$ and are zero outside the respective element. The global approximation $u^\delta$ of the exact solution is obtained by summing these polynomials piecewise, resulting in a generally discontinuous function between elements.
Similarly, the exact flux $f(u)$ in eq. (\ref{1D scalar conservation law}) can be approximated using fluxes $f_j^\delta$, which are polynomials of degree $P+1$ within each $\Omega_j$ and are zero outside the element. The global approximation $f^\delta$ is obtained by summing these polynomials piecewise, resulting in a $C^0$ continuous function between elements. It is crucial for $f^\delta$ to maintain $C^0$ continuity between elements to ensure that the scheme remains conservative.

For each cell, there is a conformal spatial mapping $\mathcal{M}: {x} \rightarrow {r}$ that can be applied to transform the physical coordinates ${x}$ into standard coordinates ${r}$, which are more convenient for the cell-local coordinates. In this way, we transform each element $\Omega_j$ into a standard element $\Omega_{j s}=\{r \mid-1<r<1\}$:

\begin{equation} 
\begin{aligned}
r & =\Gamma_j(x)=2\left(\frac{x-x_j}{x_{j+1}-x_j}\right)-1, \\
x & =\Gamma_j(r)^{-1}=\left(\frac{1-r}{2}\right) x_j+\left(\frac{1+r}{2}\right) x_{j+1}.
\end{aligned}
\end{equation}

After applying these changes, eq. (\ref{1D scalar conservation law}) for the $j^{th}$ element , $\Omega_{j s}$, has the form:

\begin{equation}
\label{transformed}
\frac{\partial \hat{u}_j^\delta}{\partial t}+\frac{1}{J_j} \frac{\partial \hat{f}_j^\delta}{\partial r}=0,
\end{equation} 
where $\hat{u}_j$ is the transformed conserved scalar quantity $\hat{u}_j^\delta=u_j^\delta\left(\Gamma_j^{-1}(r), t\right)$, $\hat{f}_j$ is the transformed flux $\hat{f}_j^\delta=f_j^\delta\left(\Gamma_j^{-1}(r), t\right)$ and 
$J_j$ is the determinant of the element Jacobian, $J_j=\frac{1}{2}\left(x_{j+1}-x_j\right)$.

    The discretization of eq. (\ref{transformed}) is performed within each element by expressing the discontinuous solution $\hat{u}_j^\delta$ and the associated discontinuous flux $\hat{f}_j^{\delta D}$ as expansions using Lagrange polynomials defined on a collection of $P+1$ interior solution points ---Gauss points. Both the discontinuous solution and the flux are polynomials of degree $P$:
    
\begin{equation}
\label{eq:u}
\begin{aligned}
\hat{u}_j^\delta(r) & =\sum_{n=1}^{P+1} \hat{u}_{j n}^\delta l_n (r), \\
\end{aligned}
\end{equation}
\begin{equation}
\begin{aligned}
\hat{f}_j^{\delta D}(r) & =\sum_{n=1}^{P+1} \hat{f}_{j n}^{\delta D} l_n (r),
\end{aligned}
\end{equation}
where $\hat{u}_{j n}^\delta$ and $\hat{f}_{j n}^{\delta D}$ are the known solution and flux values respectively at interior solution points, $r_n$, and $l_n$ are the Lagrange polynomials.

To establish common interface fluxes, the discontinuous solution values are computed at the element interfaces using eq. (\ref{eq:u}). The corresponding interface values from neighboring elements are then employed as the left and right states within a suitable numerical flux formulation for the equation being solved:

$$
\begin{aligned}
\hat{f}_j^{\delta L} & =F\left(\hat{u}_{j-1}^\delta(1), \hat{u}_j^\delta(-1)\right), \\
\hat{f}_j^{\delta R} & =F\left(\hat{u}_j^\delta(1), \hat{u}_{j+1}^\delta(-1)\right),
\end{aligned}
$$
where $F(L, R)$ is the flux function at the interface and $\hat{f}_j^{\delta L}$ and $\hat{f}_j^{\delta R}$ are the common interface fluxes on the left and right boundaries of the $j^{th}$ element, respectively.
In this work, we use the Roe-type Riemann solver for the inviscid interface fluxes and a local discontinuous Galerkin Riemann solver for the viscous interface fluxes.

To achieve a globally $C^0$ continuous flux $\hat{f}^\delta$ that aligns with the common interface flux values at element interfaces, we construct a continuous flux $\hat{f}_j^\delta$ by incorporating correction functions $g_L(r)$ and $g_R(r)$ into the discontinuous flux $\hat{f}_j^{\delta D}$. These correction functions are polynomials of degree $P+1$ and are designed in such a way that $g_L(r)$ takes the value of 1 at the left boundary, 0 at the right boundary, and 0 at all the solution points. Similarly, $g_R(r)$ is defined with the boundaries swapped. A widely adopted approach for correction functions, $g$, is to utilize a combination of the left- and right-Radau polynomials.

The continuous flux is constructed by forming a correction flux
\begin{equation}
\hat{f}_j^{\delta C}(r)=\Delta \hat{f}_L g_L(r)+\Delta \hat{f}_R g_R(r),
\end{equation}
and adding it to the existing discontinuous flux
\begin{equation}
\hat{f}_j^\delta=\hat{f}_j^{\delta D}(r)+\hat{f}_j^{\delta C}(r),
\end{equation}
with
\begin{equation}
\begin{aligned}
& \Delta \hat{f}_L=\hat{f}_j^{\delta L}-\hat{f}_j^{\delta D}(-1), \\
& \Delta \hat{f}_R=\hat{f}_j^{\delta R}-\hat{f}_j^{\delta D}(1),
\end{aligned}
\end{equation}
where $\hat{f}_j^{\delta C}$ is the correction flux (a polynomial of degree $P+1$) and $\hat{f}_j^\delta$ is the resulting continuous flux (a polynomial of degree $P+1$), which takes the values of the common interface fluxes at the boundaries.

Subsequently, the continuous flux $\hat{f}_j^\delta$ obtained is employed in eq. (\ref{transformed}) along with a time marching scheme, which in this study is the Runge-Kutta 4 method. This combination is used to update the solution within each element:

\begin{equation}
\label{eq: fluxx}
\frac{\partial u_j^\delta}{\partial t}=-\frac{1}{J_j} \frac{\partial}{\partial r}\left[f_j^{\delta D}(r)+f_j^{\delta C}(r)\right].
\end{equation}

More details can be found in \cite{Castonguay,WITHERDEN2016227}.

\subsection{Multigrid methods}
Numerous iterative solvers applied to partial differential equations reduce high-frequency components efficiently; while low-frequency components ---a smooth error--- converge more slowly \cite{Hirsch-1988,Multigrid}.
Multigrid methods aim for solving this well-known problem by successively coarsening the mesh in a hierarchical way, and then reconstructing the solution in the original fine mesh.

Depending on the way the coarsening mesh hierarchy is defined, we can find two categories: h-multigrid and p-multigrid \cite{JOMO2021114075}. While h-multigrid is based on a hierarchy of increasingly coarser geometric meshes by merging cells; the hierarchy in p-multigrid is constructed by reducing the polynomial degree of the shape functions in each multigrid level.
A convenient approach to maximize the advantages of multigrid in high-order methods is a combination of h/p-multigrid, which consists in implementing p-multigrid and then adding h-multigrid levels in a combined hp-hierarchy (or vice versa) \cite{hp-multigrid,NASTASE2006330}.

\begin{figure}
  \centering




\includegraphics[width=0.8\textwidth]{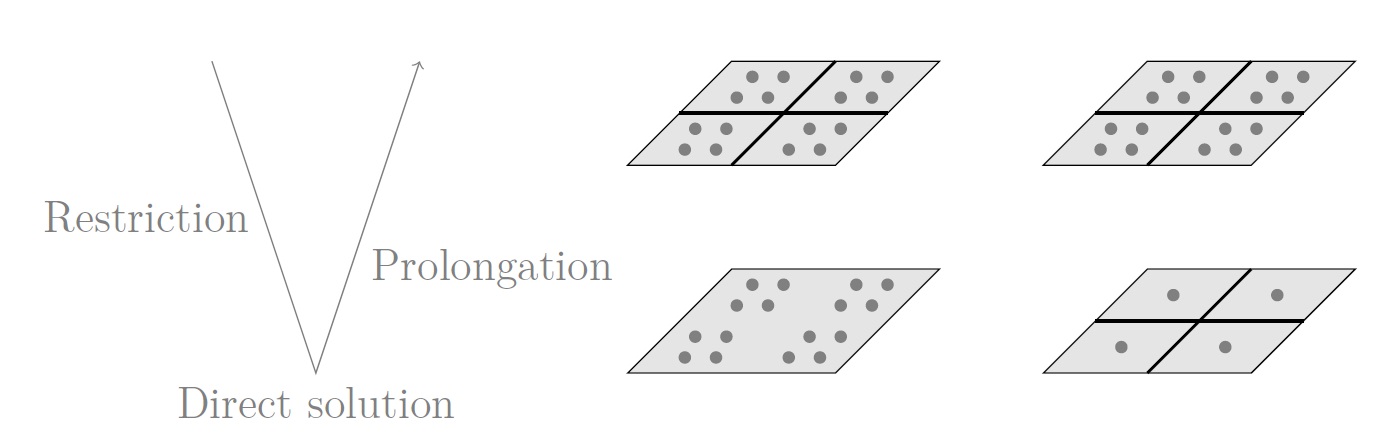}
\caption{Scheme of a V-cycle (left) of a one level h-multigrid (center) and p-multigrid
(right) \cite{Antonietti2018}. The number of degrees of freedom are represented by dots.}
\label{fig:hp}
\end{figure}

Regardless of the specific multigrid type or whether it is being used for a linear or nonlinear problem, all multigrid solvers have three essential components:

\begin{itemize}
\item \textit{Smoothing} or \textit{relaxation}: This component is responsible for ``smoothing'' the error on any given level. In most cases, it involves another iterative solver embedded in the multigrid (e.g. a Runge-Kutta scheme). Relaxation strategies can differ between different multigrid levels.
\item \textit{Coarse grid transfer}: This involves the interpolation of different variables between different multigrid levels. Transfer from a fine grid to a coarse grid is commonly known as \textit{restriction}, while the inverse operation is referred to as \textit{prolongation}. Universal transfer techniques on simple and complex geometries are covered in \cite{Brandt1984,Trottenberg2001}. To increase effectiveness, transfer operators are usually tailored for a particular discretization method and application.
\item \textit{Cycling strategy}: This component covers other parameters, including the number of coarse grids that are used and the scheduling of when they are transferred. The primary cycling strategy is known as the \textit{V-cycle}: the error is relaxed on the finest level, and then, level by level, onto the coarsest grid and back again onto the finest grid. For illustration, Figure \ref{fig:hp} depicts one V-cycle when combining the h and p levels. Other popular choices are the \textit{w-cycle} and \textit{full multigrid} (FMG) \cite{Brandt1984,Trottenberg2001}. In this work we focus on V-cycle strategies, but the presented methodology can be extended to other cycling strategies.
\end{itemize}

In this work, eq. \eqref{eq: fluxx} is accelerated to steady-state conditions using a combined h- and p-multigrid hierarchy with Runge-Kutta 4 as a smoother. We will distinguish between cases where only the p-multigrid is used and where combinations of h/p are selected, and we will study the efficiency of RL in accelerating these combinations.
%
%
The final acceleration that can be achieved depends on two essential parameters that define multigrid methods:
the correction fraction and the number of smoothing sweeps.
\begin{itemize}
\item \textit{Correction Fraction ($\alpha$):} The correction fraction, denoted by $\alpha$, represents the proportion of the corrected solution that is interpolated from a coarser grid to a finer grid. A lower correction fraction implies increased stability for the multigrid method.

\item \textit{Smoothing Sweeps:} Smoothing sweeps refer to the number of iterations performed in each level of the V-cycle to smooth out the error in the approximated solution, effectively reducing its high-frequency error. The required number of smoothing sweeps depends on the specific equation being solved and the chosen smoother. Smoothing sweeps are categorized into pre-smoothing sweeps and post-smoothing sweeps, conducted before and after the coarsest grid solution step, respectively. In practice, the number of pre-smoothing sweeps must be sufficiently high to ensure the effective smoothing of high-frequency error modes. This step is vital to prevent the introduction of noise into lower multigrid levels by the restriction operation. Similarly, the number of post-smoothing sweeps should be adequate to prevent mid-frequency error modes from developing into higher-order representations after the prolongation and the correction.
\end{itemize}

More details can be found in \cite{joshi2023robust}.  

\subsection{Summary of the numerical method to be optimized}
In the h/p-multigrid framework, the p-multigrid is performed first, followed by the h-multigrid.
The p-multigrid algorithm decreases the polynomial order at each p-level until it reaches order zero.  Then,
the h-multigrid is applied with three coarsening h-levels, and finally the solution is prolonged
and corrected.
Considering the number of sweeps, in our solver there is a linear sweep pattern; that is, the number of sweeps performed in each level is linearly increased:
one sweep is performed in the coarsest level, two sweeps in the next finer level, ..., until the finest level is reached. 


\subsection{Proximal Policy Optimization (PPO)}

 The core idea of this work is to use reinforcement learning to automatically select the multigrid parameters as the solution evolves in time. Note that these parameters may change as the solution advances, and hence the optimization needs to be dynamic. We train the PPO agent------which consists of two neural networks--- to learn an optimal policy that guides the p-multigrid process by dynamically choosing its optimal parameters, i.e. the number of smoothing sweeps at each p-level and the correction fraction. 
By incorporating the knowledge acquired through training, the agent can then exploit the hierarchical structure of the multigrid method to efficiently navigate the solution space in a wide variety of situations.

Within the reinforcement learning framework, the flux-reconstruction solver serves as the environment.
Additionally, state tuples and a reward function must be defined to inform the agent about the solver's performance and the effectiveness of each action. Furthermore, the reward function can be designed to encourage faster convergence, reduce computational effort, or other desired characteristics. 
With all these elements in mind, the PPO algorithm updates the weights of the neural networks to maximize the expected cumulative rewards, i.e. minimize the losses of the networks with a gradient descent algorithm \cite{Schulman}.
Through iterations of training and policy updates, the agent progressively improves its decision-making strategy and learns to exploit the p-multigrid hierarchy effectively. As a result, the PPO-based approach can significantly reduce the run-time until convergence is reached, thereby accelerating the solution process for partial differential equations.

In subsequent sections, we will explore the integration of PPO with multigrid methods for solving advection-diffusion equations. Also, we will investigate the impact of the PPO-based approach in the p-multigrid compared to a traditional multigrid method.

\subsection{PPO framework}
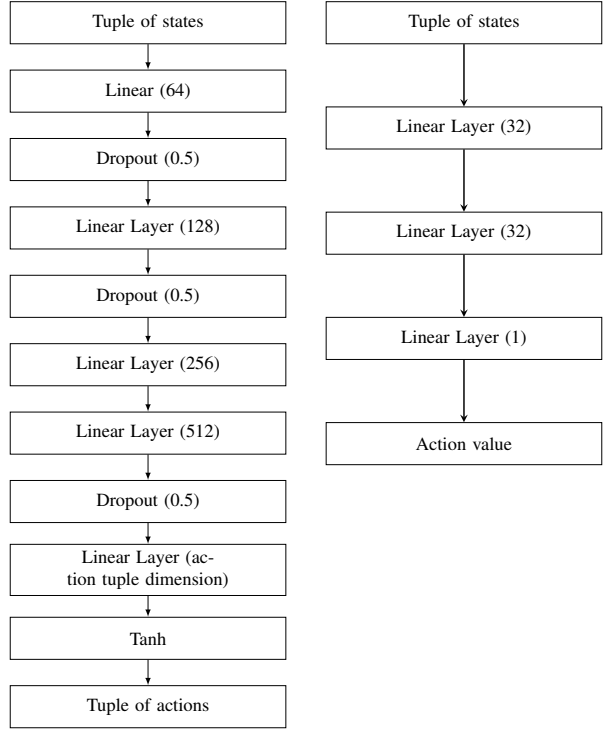
\begin{figure}
\centering
\begin{adjustbox}{scale=0.7}
\begin{tikzpicture}[node distance=1.3cm, block/.style={rectangle, draw, fill=white, text width=5cm, text centered, minimum height=0.8cm}, line/.style={draw, -latex}, arrow/.style={thick,->,>=stealth}]

\node [block] (input1) {Tuple of states};
\node [block, below of=input1] (fc11) {Linear (64)};
\node [block, below of=fc11] (dropout11) {Dropout (0.5)};
\node [block, below of=dropout11] (fc21) {Linear Layer (128)};
\node [block, below of=fc21] (dropout21) {Dropout (0.5)};
\node [block, below of=dropout21] (fc31) {Linear Layer (256)};
\node [block, below of=fc31] (fc41) {Linear Layer (512)};
\node [block, below of=fc41] (dropout41) {Dropout (0.5)};
\node [block, below of=dropout41] (fc51) {Linear Layer (action tuple dimension)};
\node [block, below of=fc51] (tanh1) {Tanh};
\node [block, below of=tanh1] (output1) {Tuple of actions}; 

\path [line] (input1) -- (fc11);
\path [line] (fc11) -- (dropout11);
\path [line] (dropout11) -- (fc21);
\path [line] (fc21) -- (dropout21);
\path [line] (dropout21) -- (fc31);
\path [line] (fc31) -- (fc41);
\path [line] (fc41) -- (dropout41);
\path [line] (dropout41) -- (fc51);
\path [line] (fc51) -- (tanh1);
\path [line] (tanh1) -- (output1);
\begin{scope}[node distance=2cm, xshift=6cm]
\node [block] (input2) {Tuple of states};
\node [block, below of=input2] (layer12) {Linear Layer (32)};
\node [block, below of=layer12] (layer22) {Linear Layer (32)};
\node [block, below of=layer22] (layer32) {Linear Layer (1)};
\node [block, below of=layer32] (output2) {Action value};

\draw [arrow] (input2) -- (layer12);
\draw [arrow] (layer12) -- (layer22);
\draw [arrow] (layer22) -- (layer32);
\draw [arrow] (layer32) -- (output2);
\end{scope}
\end{tikzpicture}
\end{adjustbox}
\caption[PPO-Clip agent structure implemented.]{Actor (left) and critic (right) neural networks architectures used in the implemented PPO. Each box shows the type of layer, followed by the number of neurons or the proportion of dropout in parentheses. The arrows show the direction in which the information is sent. ReLU is used as the activation function for all the layers.}
\label{fig:PPO-NN}
\end{figure}

Considering the previous explanation, the choices for the reinforcement elements are the following:

\begin{itemize}
    \item{Environment:} The environment is a one-dimensional flux-reconstruction solver with a h/p-multigrid algorithm; although the PPO agent will only modify the p-multigrid part. Its main features are summarized in Table \ref{tab:Params}. 

    \begin{table}[ht]
  \centering
  \caption[Numerical methods implemented in the CFD solver.]{Numerical methods implemented in the solver to be optimized.}
  \renewcommand{\arraystretch}{1.5}
  \label{tab:Params}
  \small
  \begin{tabular}{|l|c|}
  \hline
    \cellcolor{gray!10} High-order method & Flux-reconstruction \\
    \hline
    \cellcolor{gray!10} Time marcher & Runge-Kutta 4 \\
    \hline
   \cellcolor{gray!10} Accelerator & h/p-multigrid \\
    \hline
    \cellcolor{gray!10} Integration points & Gauss \\
    \hline
    \cellcolor{gray!10} Integration polynomial & Lagrange \\
    \hline
  \end{tabular}
\end{table}
    
    \item{Reward:}  It is the objective function to be optimized during the training. Given that the purpose of coupling reinforcement learning with the flux-reconstruction solver is to efficiently foster convergence, a big relative drop in the residual and a short time to convergence are desired. 
    Consequently, the reward at each step designed for this problem is: 
    \begin{equation}
            R_t=\frac{r_{t-1}-r_t}{r_{t-1} \Delta t_{\text {clock }}},
        \end{equation}
    where $\frac{r_{t-1}-r_t}{r_{t-1}}$ is the relative drop in the residual at that step ---note that if it diverges, the reward will be negative, becoming a ``punishment'' for the agent--- and $\Delta t_{\text {clock}}$ is the time taken for that operation ---the larger it is, the lower the reward for the action taken.

    \item{State:} It is the information from the environment that is provided to the agent, so that it can make decisions in consequence. One of the main characteristics that defines the process of solving a partial differential equation is the relative drop in the residual at a step $t$, which is also characterized by its sign: positive when converging; and negative if diverging. Consequently, in the state tuple, the relative drop in the residual $\frac{r_{t-1}-r_t}{r_{t-1}}$ must be taken into account. 
    Furthermore, through multiple training and testing trials, it was observed that explicitly including its sign facilitated the training process. As a result, this additional information was incorporated into the state. 
    Finally, it was observed in the original solver that the run-time increased as the equations' coefficients changed (e.g. the advection velocity and the kinematic viscosity in eq. (\ref{eq: ad-dif})). Consequently, those coefficients were added to the state tuple. Note that during the training process those coefficients will be continuously changed after a certain number of episodes, while during the tests, they are kept as constant.
    Therefore, the final state tuple is: [ $\frac{r_{t-1}-r_t}{r_{t-1}}$,  $sign (\frac{r_{t-1}-r_t}{r_{t-1}}$), $\textit{eq. coefficients}$].

    \item{Actions:} They are the decisions that the agent can make based on the information from the state. As the purpose of implementing PPO is to accelerate convergence, the actions are defined as selecting the right p-multigrid parameters, namely the correction fraction for the finest level and the number of smoothing sweeps at each p-level ---divided into pre-smoothing sweeps and post-smoothing sweeps, depending if they are performed before or after the coarsest level solution, respectively. Consequently, the action tuple is defined as:  
    {$[{\textit{Pre-smoothing sweeps}},\, {\textit{Post-smoothing sweeps}},\, {\textit{Correction fraction}}].$} \\
    It is important to notice that, as $P$ is the maximum polynomial order used to solve the PDE and $0$ is the minimum, 
    the pre-sweeps tuple will be $P$-dimensional and the post-sweeps tuple, $P-1$-dimensional. 
    
    
     \item{Agent:} The PPO agent, that receives the state tuple as input and generate the appropriate actions, is divided into two neural networks: the actor ---which will choose the action tuple to be applied in the next step--- and the critic ---which evaluates the decision made by the actor. Only the actor will interact with the p-multigrid once trained. Regarding their structure, both are feed forward neural networks \cite{SURESHKUMAR2020100288}.
     The scheme of both actor and critic neural networks is shown in Figure \ref{fig:PPO-NN}. This specific structure was chosen after observing several trainings with different neural network architectures. 
    
\end{itemize}

The hyperparameters used in our PPO have been manually selected after multiple trial and error, and they are shown in Table \ref{tab: PPO Hyper}. Note that one step in each episode is one complete V-cycle for which the p-multigrid parameters are adjusted by the PPO agent. We have found that the PPO is not very sensitive to hyperparameters and does not
need extensive optimization, while the most relevant is learning rate, initially set to $0.05$, but later changed to $10^{-5}$ (for the actor NN), as it was observed that smaller policy updates
during training are more likely to converge. 

\begin{table}[ht]
  \centering
    \small
    \renewcommand{\arraystretch}{0.8}
  \caption[PPO-clip hyperparameters]{PPO hyperparameters.}
  \renewcommand{\arraystretch}{2} 
  \label{tab:PPO Hyper}
  \begin{tabular}{|l|c|}
    \hline
    \cellcolor{gray!10} Nº of training episodes & 10,000 \\
    \hline
    \cellcolor{gray!10} Timesteps per batch & 16 \\
    \hline
   \cellcolor{gray!10} Discount factor $\gamma$ & 0.98 \\
    \hline
    \cellcolor{gray!10} Actor learning rate & $10^{-5}$ \\
    \hline
    \cellcolor{gray!10} Critic learning rate & 0.05 \\
    \hline
    \cellcolor{gray!10} Exploration factor $\varepsilon$ & 0.2 \\
        \hline
  \end{tabular}
\label{tab: PPO Hyper}
\end{table}

The previous hyperparameters include:
\begin{itemize}
\item Discount factor $\gamma$: hyperparameter that adjusts the importance of rewards
over time, generally fixed to a high value between 0 and 1. The discount
factor determines how much the agent cares about rewards in the long term.
\item The exploration factor $\varepsilon$: determines the balance between exploration and
exploitation and it is bounded between 0 and 1. During the training, the agent occasionally performs random exploration
 with probability $\varepsilon$ and takes the optimal action with probability $1-\varepsilon$.
\end{itemize}
     
\section{Results}

In this section, we present the results obtained from several tests after training the PPO agent, comparing its performance with that of the original solver without RL. We aim to evaluate two aspects: first, how well the RL strategy performs against heuristically selected parameters, and second, how the performance is affected depending on the way the PPO agent has been trained.
For the runs in the original solver, the number of pre- and post-smoothing sweeps was chosen to follow a linear progression across multigrid levels, and the correction fraction was set to 0.1 for the finest level and 1 for the rest.
The p-multigrid is set so that the polynomial order is decreased one unit per level, until $P=0$ is reached at the coarsest level.
The h-multigrid part ---always used in a h/p-multigrid context--- will be applied after the p-multigrid with three coarser levels. These levels are defined by halving the number of elements in the mesh at each level. Finally, all test cases are converged until the residual reaches $10^{-9}$.
\\

We first study the effect of using a uniform mesh in the training process. We train two PPO agents: one in a uniform mesh environment and the other in a non-uniform mesh environment, both within an h/p-multigrid context. The polynomial order used in the FR scheme remains constant with a value of $P=2$.
Table \ref{tab:uni vs nonuni} presents a comparison of the performance between the original solver, the PPO agent trained in a uniform mesh, and the PPO agent trained in a non-uniform mesh environment, to solve equations in a uniform mesh setting. The PPO agent trained in a non-uniform mesh has shown comparable performance to its uniform mesh counterpart.  However, when tested in a non-uniform mesh environment, the PPO agent trained in a uniform mesh consistently diverged, indicating its inability to solve the equations under these conditions.
Both PPO agents maintain a consistent number of iterations and run-time across various tested parameter values (advection speed $a$ and viscosity $\nu$). In contrast, the heuristic approach shows an increase in both the number of iterations and the execution time as the viscosity ($\nu$) is increased.

Once the superiority of non-uniform mesh training over uniform mesh training has been proven, we evaluate the performance of the PPO agent trained in a non-uniform mesh within an h/p-multigrid environment for solving problems with non-uniform meshes. Table \ref{tab:hp} compares the performance of this PPO agent with that of the original solver in an h/p-multigrid context with varying parameters ($a$ and $\nu$). The polynomial order used in the FR scheme ranges from 2 to 5.  The results illustrate significant improvements in convergence times by factors of up to more than 100 in the tested cases. Again, the PPO agent maintains consistent run-time and iteration counts, while these variables increase with heuristically chosen parameters.

Finally, we evaluate the importance of training in an h/p-multigrid environment. We trained two PPO agents under different conditions: one under h/p-multigrid and the other under p-multigrid. Their performance was compared to that of the original solver in a p-multigrid context, where no h-multigrid levels were applied afterwards. The polynomial order used in the FR scheme ranges from 2 to 3. The results are presented in Table \ref{tab: p}, which shows that both variants of PPO demonstrate similar efficiency, again keeping consistent run-time and iteration counts.
In contrast, we also tested the reverse scenario and found that the PPO agent trained in a p-multigrid context consistently failed to solve the equations in an h/p-multigrid environment, resulting in solution divergence.\\ 

Based on the previous results, several observations were noted:

\begin{itemize}
    \item[$-$ ]Proximal Policy Optimization (PPO) is a successful method to foster convergence in h/p-multigrid by dynamically choosing the number of sweeps at each multigrid level and the correction fraction before the finest level in the p-multigrid part. The method is efficient for a large range of parameters (advection speed and viscosity), poolynomial orders and mesh spacing. 
    \item[$-$ ] It is beneficial to train the most complex scenario that will be found by the agent in a real problem. In particular, a trained PPO in an h/p-multigrid environment works correctly in a p-multigrid context when tested. On the contrary, a trained PPO in a p-multigrid environment is not applicable in a h/p-multigrid context. Also, a trained PPO in a non-uniform mesh environment works correctly in an uniform mesh context when tested. On the contrary, a trained PPO in an uniform mesh environment is not applicable in a non-uniform mesh context.
    \item[$-$ ] When using a linear sweep pattern, run-time varies with the equation coefficients. On the contrary, when optimizing the solving process with PPO, the run-times are more consistent regardless of the coefficients of the equation being solved.    
\end{itemize}

\begin{adjustwidth}{-2.5 cm}{-2.5 cm}
\centering\begin{threeparttable}[!htb]
\caption[Original runs compared to PPO runs -trained for uniform and non-uniform meshes- for order 2 in a uniform mesh.]{Run-time and number of iterations of the runs of the original solver compared to PPO ---trained in uniform and non-uniform meshes environments--- for  $P=2$ in a h/p-multigrid, uniform mesh context. For clarity, shortest run-times are marked in bold.}\label{tab:uni vs nonuni}
\scriptsize
\renewcommand{\arraystretch}{1.3}
\begin{tabular}{cc|cc|cc}
\multicolumn{6}{c}{\cellcolor[HTML]{BFBFBF}\textbf{h/p-multigrid, uniform mesh}} \\
\multicolumn{2}{c|}{\cellcolor[HTML]{BFBFBF}\textbf{Original}} &\multicolumn{2}{c|}{\cellcolor[HTML]{BFBFBF}\textbf{PPO (Uniform)}} &\multicolumn{2}{c}{\cellcolor[HTML]{BFBFBF}\textbf{PPO (Non-uniform)}} \\
\cellcolor[HTML]{BFBFBF}\textbf{Runtime (s)} &\cellcolor[HTML]{BFBFBF}\textbf{Nº iterations} &\cellcolor[HTML]{BFBFBF}\textbf{Runtime (s)} &\cellcolor[HTML]{BFBFBF}\textbf{Nº iterations} &\cellcolor[HTML]{BFBFBF}\textbf{Runtime (s)} &\cellcolor[HTML]{BFBFBF}\textbf{Nº iterations} \\
\multicolumn{6}{c}{\cellcolor[HTML]{BFBFBF}\textbf{Linear Advection-Diffusion - $P=2$}} \\
\multicolumn{6}{c}{\cellcolor[HTML]{E3E3E3}a= 1.0, $\nu$ = 0.01} \\
12.42 &83 &3.91 &104 &\textbf{3.74} &104 \\
\multicolumn{6}{c}{\cellcolor[HTML]{E3E3E3}a= 0.5, $\nu$ = 0.01} \\
16.29 &111 &3.96 &104 &\textbf{3.76} &105 \\
\multicolumn{6}{c}{\cellcolor[HTML]{E3E3E3}a= 0.5, $\nu$ = 0.5} \\
60.33 &374 &3.91 &104 &\textbf{3.74} &105 \\
\multicolumn{6}{c}{\cellcolor[HTML]{E3E3E3}a= 0.4, $\nu$ = 0.6} \\
91.60 &552 &3.85 &104 &\textbf{3.74} &106 \\
\multicolumn{6}{c}{\cellcolor[HTML]{E3E3E3}a= 0.2, $\nu$ = 0.8} \\
219.45 &1427 &3.89 &105 &\textbf{3.75} &104 \\
\multicolumn{6}{c}{\cellcolor[HTML]{BFBFBF}\textbf{Burgers - $P=2$}} \\
\multicolumn{6}{c}{\cellcolor[HTML]{E3E3E3}$\nu$ = 0.05} \\
193.68 &655 &5.34 &133 &\textbf{5.03} &130 \\
\multicolumn{6}{c}{\cellcolor[HTML]{E3E3E3}$\nu$ = 0.3} \\
109.10 &363 &5.43 &134 &\textbf{4.87} &129 \\
\multicolumn{6}{c}{\cellcolor[HTML]{E3E3E3} $\nu$ = 0.5} \\
206.65 &705 &5.30 &131 &\textbf{5.13} &132 \\
\multicolumn{6}{c}{\cellcolor[HTML]{E3E3E3}$\nu$ = 0.8} \\
323.25 &1240 &5.22 &130 &\textbf{4.65} &124 \\
\multicolumn{6}{c}{\cellcolor[HTML]{E3E3E3}$\nu$ = 1.0} \\
356.03 &1602 &5.12 &132 &\textbf{5.06} &133 \\
\end{tabular}
\end{threeparttable}
\end{adjustwidth}

\begin{table}[!htp]
\centering
\caption{Run-time and number of iterations of the runs of the original solver compared to PPO for $P=2,3,5$ in a h/p-multigrid, non-uniform mesh context. For clarity, shortest run-times are marked in bold.}
\label{tab:hp}
\scriptsize
\begin{tabular}{cc|cc}
\multicolumn{4}{c}{\cellcolor[HTML]{BFBFBF} \textbf{h/p-multigrid, non-uniform mesh}} \\
\multicolumn{2}{c|}{\cellcolor[HTML]{BFBFBF} \textbf{Original}} &\multicolumn{2}{c}{\cellcolor[HTML]{BFBFBF} \textbf{PPO}} \\
\cellcolor[HTML]{BFBFBF} \textbf{Runtime (s)} &\cellcolor[HTML]{BFBFBF} \textbf{Nº iterations} &\cellcolor[HTML]{BFBFBF} \textbf{Runtime (s)} &\cellcolor[HTML]{BFBFBF} \textbf{Nº iterations} \\
\multicolumn{4}{c}{\cellcolor[HTML]{BFBFBF} \textbf{Linear advection-diffusion - $P=2$}} \\
\multicolumn{4}{c}{\cellcolor[HTML]{E3E3E3}a= 1.0, $\nu$ = 0.01} \\
69.72 &197 &\textbf{31.03} &626 \\
\multicolumn{4}{c}{\cellcolor[HTML]{E3E3E3}a= 0.5, $\nu$ = 0.01} \\
80.01 &207 &\textbf{31.81} &651 \\
\multicolumn{4}{c}{\cellcolor[HTML]{E3E3E3}a= 0.5, $\nu$ = 0.5} \\
808.60 &2178 &\textbf{33.21} &652 \\
\multicolumn{4}{c}{\cellcolor[HTML]{E3E3E3}a= 0.4, $\nu$ = 0.6} \\
634.27 &3166 &\textbf{31.52} &654 \\
\multicolumn{4}{c}{\cellcolor[HTML]{E3E3E3}a= 0.2, $\nu$ = 0.8} \\
1,476.48 &8063 &\textbf{31.48} &648 \\
\multicolumn{4}{c}{\cellcolor[HTML]{BFBFBF} \textbf{Linear Advection-Diffusion - $P=3$}} \\
\multicolumn{4}{c}{\cellcolor[HTML]{E3E3E3}a= 1.0, $\nu$ = 0.01} \\
51.73 &228 &\textbf{33.98} &658 \\
\multicolumn{4}{c}{\cellcolor[HTML]{E3E3E3}a= 0.5, $\nu$ = 0.01} \\
54.73 &237 &\textbf{35.34} &659 \\
\multicolumn{4}{c}{\cellcolor[HTML]{E3E3E3}a= 0.5, $\nu$ = 0.5} \\
2,168.68 &6094 &\textbf{130.78} &1602 \\
\multicolumn{4}{c}{\cellcolor[HTML]{E3E3E3}a= 0.4, $\nu$ = 0.6} \\
1,861.64 &8915 &\textbf{178.78} &2201 \\
\multicolumn{4}{c}{\cellcolor[HTML]{E3E3E3}a= 0.2, $\nu$ = 0.8} \\
2,467.52 &9444 &\textbf{157.62} &1958 \\
\multicolumn{4}{c}{\cellcolor[HTML]{BFBFBF} \textbf{Linear Advection-Diffusion - $P=5$}} \\
\multicolumn{4}{c}{\cellcolor[HTML]{E3E3E3}a= 1.0, $\nu$ = 0.01} \\
\textbf{142.62} &479 &1,081.12 &7094 \\
\multicolumn{4}{c}{\cellcolor[HTML]{E3E3E3}a= 0.5, $\nu$ = 0.01} \\
2,500.01 &8053 &\textbf{1,203.32} &7126 \\
\multicolumn{4}{c}{\cellcolor[HTML]{E3E3E3}a= 0.5, $\nu$ = 0.5} \\
17,976.47 &37647 &\textbf{1,200.36} &7102 \\
\multicolumn{4}{c}{\cellcolor[HTML]{E3E3E3}a= 0.4, $\nu$ = 0.6} \\
18,003.09 &37647 &\textbf{1,213.74} &7115 \\
\multicolumn{4}{c}{\cellcolor[HTML]{E3E3E3}a= 0.2, $\nu$ = 0.8} \\
2,467.52 & 9444 &\textbf{1,194.44} &7101 \\
\multicolumn{4}{c}{\cellcolor[HTML]{BFBFBF} \textbf{Burgers - $P=2$}} \\
\multicolumn{4}{c}{\cellcolor[HTML]{E3E3E3}$\nu$ = 0.05} \\
296.45 &799 &\textbf{90.81} &1524 \\
\multicolumn{4}{c}{\cellcolor[HTML]{E3E3E3}$\nu$ = 0.3} \\
1,022.65 &3109 &\textbf{79.14} &1328 \\
\multicolumn{4}{c}{\cellcolor[HTML]{E3E3E3} $\nu$ = 0.5} \\
1,593.30 &5222 &\textbf{82.47} &1388 \\
\multicolumn{4}{c}{\cellcolor[HTML]{E3E3E3}$\nu$ = 0.8} \\
2,544.08 &8853 &\textbf{84.51} &1400 \\
\multicolumn{4}{c}{\cellcolor[HTML]{E3E3E3}$\nu$ = 1.0} \\
3,102.94 &11209 &\textbf{99.00} &1644 \\
\multicolumn{4}{c}{\cellcolor[HTML]{BFBFBF} \textbf{Burgers - $P=3$}} \\
\multicolumn{4}{c}{\cellcolor[HTML]{E3E3E3}$\nu$ = 0.05} \\
453.92 &1351 &\textbf{84.66} &1405 \\
\multicolumn{4}{c}{\cellcolor[HTML]{E3E3E3}$\nu$ = 0.3} \\
2,729.82 &7371 &\textbf{86.48} &1411 \\
\multicolumn{4}{c}{\cellcolor[HTML]{E3E3E3} $\nu$ = 0.5} \\
4,990.60 &14402 &\textbf{86.03} &1400 \\
\multicolumn{4}{c}{\cellcolor[HTML]{E3E3E3}$\nu$ = 0.8} \\
10,067.14 &31683 &\textbf{94.25} &1534 \\
\multicolumn{4}{c}{\cellcolor[HTML]{E3E3E3}$\nu$ = 1.0} \\
13,590.96 &47423 &\textbf{74.06} &1217 \\
\multicolumn{4}{c}{\cellcolor[HTML]{BFBFBF} \textbf{Burgers - $P=5$}} \\
\multicolumn{4}{c}{\cellcolor[HTML]{E3E3E3}$\nu$ = 0.05} \\
1,290.16 &3681 &\textbf{830.71} &7621 \\
\multicolumn{4}{c}{\cellcolor[HTML]{E3E3E3}$\nu$ = 0.3} \\
8,566.44 &21486 &\textbf{1,602.69} &7680 \\
\multicolumn{4}{c}{\cellcolor[HTML]{E3E3E3} $\nu$ = 0.5} \\
4,990.60 & 14402 &\textbf{1,536.88} &7343 \\
\multicolumn{4}{c}{\cellcolor[HTML]{E3E3E3}$\nu$ = 0.8} \\
10,067.14 & 31683 &\textbf{1,545.23} &7573 \\
\multicolumn{4}{c}{\cellcolor[HTML]{E3E3E3}$\nu$ = 1.0} \\
13,590.96 & 47423 &\textbf{1,640.68} &8007 \\
\end{tabular}
\end{table}

\begin{adjustwidth}{-2.5cm}{-2.5cm}
\centering
\begin{threeparttable}[!htb]
\caption{Run-time and number of iterations of the runs of the original solver compared to PPO ---trained for p-multigrid and h/p-multigrid--- for $P=2,3$ in a p-multigrid, non-uniform mesh context. For clarity, shortest run-times are marked in bold.}\label{tab: p}
\scriptsize
\renewcommand{\arraystretch}{1.3}
\begin{tabular}{cc|cc|cc}
\multicolumn{6}{c}{\cellcolor[HTML]{BFBFBF}\textbf{p-multigrid, non-uniform mesh}} \\
\multicolumn{2}{c|}{\cellcolor[HTML]{BFBFBF}\textbf{Original}} &\multicolumn{2}{c|}{\cellcolor[HTML]{BFBFBF}\textbf{PPO (p-multigrid)}} &\multicolumn{2}{c}{\cellcolor[HTML]{BFBFBF}\textbf{PPO (h/p-multigrid)}} \\
\cellcolor[HTML]{BFBFBF}\textbf{Runtime (s)} &\cellcolor[HTML]{BFBFBF}\textbf{Nº iterations} &\cellcolor[HTML]{BFBFBF}\textbf{Runtime (s)} &\cellcolor[HTML]{BFBFBF}\textbf{Nº iterations} &\cellcolor[HTML]{BFBFBF}\textbf{Runtime (s)} &\cellcolor[HTML]{BFBFBF}\textbf{Nº iterations} \\
\multicolumn{6}{c}{\cellcolor[HTML]{BFBFBF}\textbf{Linear Advection-Diffusion - $P=2$}} \\
\multicolumn{6}{c}{\cellcolor[HTML]{E3E3E3}a= 1.0, $\nu$ = 0.01} \\
73.56 &455 &36.35 &656 &\textbf{35.33} &652 \\
\multicolumn{6}{c}{\cellcolor[HTML]{E3E3E3}a= 0.5, $\nu$ = 0.01} \\
131.11 &710 &35.11 &658 &\textbf{34.99} &665 \\
\multicolumn{6}{c}{\cellcolor[HTML]{E3E3E3}a= 0.5, $\nu$ = 0.5} \\
862.87 &3883 &36.42 &665 &\textbf{34.72} &664 \\
\multicolumn{6}{c}{\cellcolor[HTML]{E3E3E3}a= 0.4, $\nu$ = 0.6} \\
899.34 &3902 &36.04 &657 &\textbf{35.83} &661 \\
\multicolumn{6}{c}{\cellcolor[HTML]{E3E3E3}a= 0.2, $\nu$ = 0.8} \\
1,554.69 &8138 &36.27 &655 &\textbf{34.98} &658 \\
\multicolumn{6}{c}{\cellcolor[HTML]{BFBFBF}\textbf{Linear Advection-Diffusion - $P=3$}} \\
\multicolumn{6}{c}{\cellcolor[HTML]{E3E3E3}a= 1.0, $\nu$ = 0.01} \\
\textbf{98.36} &228 &159.98 &1761 &157.35 &1777 \\
\multicolumn{6}{c}{\cellcolor[HTML]{E3E3E3}a= 0.5, $\nu$ = 0.01} \\
\textbf{101.84} &237 &179.32 &1945 &156.60 &1758 \\
\multicolumn{6}{c}{\cellcolor[HTML]{E3E3E3}a= 0.5, $\nu$ = 0.5} \\
5,764.68 &6094 &171.39 &1863 &\textbf{157.26} &1776 \\
\multicolumn{6}{c}{\cellcolor[HTML]{E3E3E3}a= 0.4, $\nu$ = 0.6} \\
6,601.37 &8915 &158.93 &1733 &\textbf{157.28} &1762 \\
\multicolumn{6}{c}{\cellcolor[HTML]{E3E3E3}a= 0.2, $\nu$ = 0.8} \\
15,056.47 &22961 &166.06 &1789 &\textbf{154.94} &1760 \\
\multicolumn{6}{c}{\cellcolor[HTML]{BFBFBF}\textbf{Burgers - $P=2$}} \\
\multicolumn{6}{c}{\cellcolor[HTML]{E3E3E3}$\nu$ = 0.05} \\
239.95 &631 &\textbf{74.66} &1133 &88.71 &1396 \\
\multicolumn{6}{c}{\cellcolor[HTML]{E3E3E3}$\nu$ = 0.3} \\
979.55 &3078 &97.04 &1434 &\textbf{88.62} &1381 \\
\multicolumn{6}{c}{\cellcolor[HTML]{E3E3E3}$\nu$ = 0.5} \\
1,574.76 &5222 &93.74 &1392 &\textbf{81.42} &1290 \\
\multicolumn{6}{c}{\cellcolor[HTML]{E3E3E3}$\nu$ = 0.8} \\
6,611.84 &8746 &93.90 &1369 &\textbf{88.54} &1384 \\
\multicolumn{6}{c}{\cellcolor[HTML]{E3E3E3}$\nu$ = 1.0} \\
7,099.67 &11071 &100.86 &1399 &\textbf{92.48} &1416 \\
\multicolumn{6}{c}{\cellcolor[HTML]{BFBFBF}\textbf{Burgers - $P=3$}} \\
\multicolumn{6}{c}{\cellcolor[HTML]{E3E3E3}$\nu$ = 0.05} \\
529.49 &1411 &\textbf{352.11} &3369 &368.75 &3341 \\
\multicolumn{6}{c}{\cellcolor[HTML]{E3E3E3}$\nu$ = 0.3} \\
3,059.29 &8448 &\textbf{346.09} &3289 &361.72 &3278 \\
\multicolumn{6}{c}{\cellcolor[HTML]{E3E3E3}$\nu$ = 0.5} \\
35,665.99 &14480 &370.69 &3307 &\textbf{363.09} &3400 \\
\multicolumn{6}{c}{\cellcolor[HTML]{E3E3E3}$\nu$ = 0.8} \\
38,188.51 &23749 &\textbf{354.75} &3389 &368.47 &3361 \\
\multicolumn{6}{c}{\cellcolor[HTML]{E3E3E3}$\nu$ = 1.0} \\
39,653.02 &29972 &343.65 &3250 &\textbf{343.44} &3133 \\
\end{tabular}
\end{threeparttable}
\end{adjustwidth}

\section{Conclusions}

In conclusion, the integration of reinforcement learning techniques with multigrid methods offers promising avenues for accelerating the solution of advection-diffusion and nonlinear Burgers' equations. Using RL (and the PPO algorithm in particular), we can enhance the parameter-related decision-making process within the multigrid solver and improve its efficiency and convergence behavior.
The use of RL enables the development of an agent that learns to exploit the hierarchical structure of the multigrid method, leading to reduced computational effort and faster convergence. The agent's policy parameters are optimized iteratively through interactions with the environment, allowing for the discovery of effective strategies for navigating the multigrid hierarchy and reducing the run-time of the simulation.
Furthermore, the integration of RL and PPO with multigrid methods opens up possibilities to address the challenges associated with complex geometries and high-order methods. RL algorithms have the potential to adapt and optimize the solver's behavior in response to varying problem setups and discretization schemes.
However, it is important to note that there are still challenges and open questions to be addressed in this field. The choice of appropriate reward formulations and generalization to different problem domains are areas that require further investigation. In addition, the scalability and applicability of the RL and PPO approaches to large-scale problems need to be thoroughly explored.


\section*{Acknowledgments}
Gonzalo Rubio and Esteban Ferrer acknowledge the funding received by the Grant DeepCFD (Project No. PID2022-137899OB-I00) funded by MICIU/AEI/10.13039/501100011033 and by ERDF, EU. 

This research has been cofunded by the European Union (ERC, Off-coustics, project number 101086075). Views and opinions expressed are, however, those of the author(s) only and do not necessarily reflect those of the European Union or the European Research Council. Neither the European Union nor the granting authority can be held responsible for them. Finally, all authors gratefully acknowledge the Universidad Politécnica de Madrid (www.upm.es) for providing computing resources on Magerit Supercomputer.

\bibliographystyle{elsarticle-harv} 
\bibliography{references}

\end{document}